%% file: main.tex
\documentclass[runningheads]{llncs}
\usepackage{graphicx}

\usepackage{tikz}
\usepackage{comment}
\usepackage{amsmath,amssymb} 
\usepackage{color}
\usepackage{hyperref}

\usepackage{xr}
\usepackage{epsfig}
\usepackage{graphicx}
\usepackage{amsmath}
\usepackage{amssymb}
\usepackage{booktabs}
\usepackage{fnpct}
\usepackage{multirow}
\usepackage[capitalize]{cleveref}
\crefname{section}{Sec.}{Secs.}
\Crefname{section}{Section}{Sections}
\usepackage[accsupp]{axessibility}  

\input{chapters/macros}

\newcommand{\etal}{\textit{et al.}}

\usepackage{wrapfig}
\newcommand\blfootnote[1]{%
  \begingroup
  \renewcommand\thefootnote{}\footnote{#1}%
  \addtocounter{footnote}{-1}%
  \endgroup
}

\begin{document}
\pagestyle{headings}
\mainmatter
\def\ECCVSubNumber{1384}  

\title{GLASS: Global to Local Attention for Scene-Text Spotting} 


\titlerunning{GLASS}
%
\author{Roi Ronen\inst{1}\thanks{Equal contribution. For correspondence: \email{tsiper@amazon.com}} \and 
Shahar Tsiper\inst{2\,\star} \and
Oron Anschel\inst{2}  \and  \\
Inbal Lavi\inst{2} \and 
Amir Markovitz\inst{2} \and
R. Manmatha\inst{2}}
\authorrunning{R. Ronen et al.}
%
\institute{Viterbi Faculty of Electrical \& Computer Eng., Technion, Haifa, Israel \and AWS AI Labs } 
\maketitle




\input{chapters/abstract_v2}
\input{chapters/intro_v2}

\input{chapters/related_work_v2}

\input{chapters/method_v2}

\input{chapters/experiments_v2}

\input{chapters/discussion}

\clearpage
%
%
\bibliographystyle{splncs04}
\bibliography{egbib}



\chapter*{Supplementary Material}

\input{supplementary/supp_content}

\end{document}

%% file: chapters/macros.tex
\ifdefined\ShowNotes
  \newcommand{\colornote}[3]{{\color{#1}\small\textsf{#2:  #3}\normalfont}}
\else
  \newcommand{\colornote}[3]{}
\fi

\newcommand {\amir}[1]{\colornote{blue}{Amir}{#1}}

%% file: chapters/abstract_v2.tex
\begin{abstract}



In recent years, the dominant paradigm for text spotting is to combine the tasks of text detection and recognition into a single \emph{end-to-end} framework. 
Under this paradigm, both tasks are accomplished by operating over a shared global feature map extracted from the input image.
Among the main challenges that end-to-end approaches face is the performance degradation when recognizing text across scale variations (smaller or larger text), and arbitrary word rotation angles.
In this work, we address these challenges by proposing a novel global-to-local attention mechanism for text spotting, termed GLASS\blfootnote{Code available at \href{https://www.github.com/amazon-research/glass-text-spotting}{{https://www.github.com/amazon-research/glass-text-spotting}}}, that fuses together global and local features.
The global features are extracted from the shared backbone, preserving contextual information from the entire image, while the local features are computed individually on resized, high resolution rotated word crops. 
The information extracted from the local crops alleviates much of the inherent difficulties with scale and word rotation.
We show a performance analysis across scales and angles, highlighting improvement over scale and angle extremities.
In addition, we introduce an orientation-aware loss term supervising the detection task, and show its contribution to both detection and recognition performance across all angles.
Finally, we show that GLASS is general by incorporating  it into other leading text spotting architectures, improving their text spotting performance.
Our method achieves state-of-the-art results on multiple benchmarks, including the newly released TextOCR.

\keywords{Text spotting, Text Detection, Text Recognition, Language Understanding.}
\end{abstract}

%% file: chapters/intro_v2.tex
\section{Introduction}
\label{sec:introduction}
\begin{figure}
 \centering
  \includegraphics[width=0.7\linewidth]{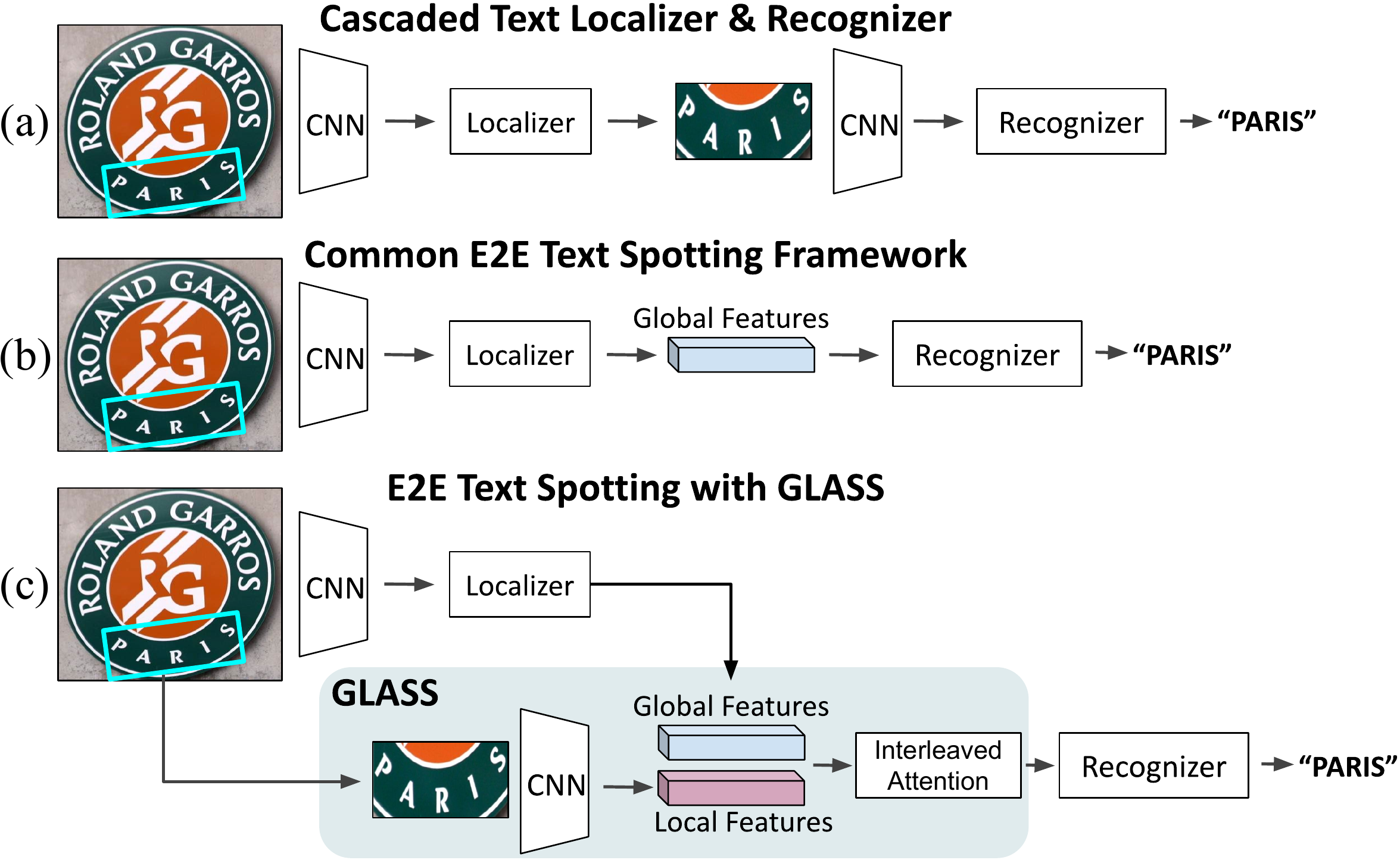}
  \caption{\textbf{An overview of text spotting approaches. }  \textbf{(a)} Cascaded. A standalone text detector followed by a standalone recognizer. Each is trained separately. 
  \textbf{(b)} End-to-end (E2E) text spotting. Detection and recognition are jointly optimized. 
  \textbf{(c)} Our approach with GLASS fusion, operating on two separate feature maps taken at different resolutions and contexts, bridging (a) and (b). Feature maps are fused using interleaved attention, improving robustness to scale and rotation, and overall performance.}
  \label{fig:main}
\end{figure}
Text spotting, the task of detecting text instances in the wild and recognizing them,  has seen a notable increase in performance in recent years.
It is now commonly used in many real-life scenarios and applications.
Demanding areas such as autonomous driving,  document analysis, and geo-localization, where accurate text transcription is a must, all rely on text spotting. 
The challenge lies in the fact that some words may span the entire image, while other words, even in the same image, may be hard to read, e.g., appear on a traffic sign barely seen from across the street. 

Two prevalent paradigms exist for text spotting (see \cref{fig:main}): the first is a modular approach, cascading independent text detection and recognition models.
The recognition model uses uniformly aligned and resized  word-crop images as input with upright orientation, abstracting away scale and rotation.
The components in this approach are mostly explored independently in the literature, isolating either the word detection performance (ignoring transcripts)~\cite{zhou2017east,buvsta2018e2e,baek2019craft,liao2020db,qin2021mask}, or the recognition performance over datasets composed of word-crop images \cite{clova,shi2018aster,litman2020scatter,lu2021master}.
The second approach is a combined End-to-End (E2E) architecture, adding a recognition branch that operates directly on the detection model's latent features~\cite{feng2019textdragon,baek2020crafts,krylov2021yamts,qiao2020mango,lyu2018spotter,liao2020spotterV3,liu2021abcnetV2}.
Feature sampling replaces cropping, allowing detection and recognition to be jointly trained E2E.

With E2E becoming the common paradigm, scale and rotation-free crops were often replaced by sampled CNN features, that are highly sensitive to both scale and rotation~\cite{Jaderberg2015SpatialTN,Worrall2017HarmonicND}.
While the joint optimization in E2E systems improved performance for average-sized and upright-facing words, scale extremities and strong rotations were overlooked.

In this work, we propose to bridge the two paradigms to get the best out of both worlds.
We combine \emph{global} features from the detector's embedding space with \emph{local} cropped word embeddings.
The fusion is done using a novel Global-to-Local interleaved attention module,
leveraging information from both feature maps.
This global-to-local approach enriches the information used by the recognition branch, boosting the overall text spotting accuracy for different scales and rotations.
Additional gains for rotated text are obtained by introducing an orientation prediction side-task, aimed at better capturing rotated words, or words in rotated images.
The side task is supervised by a new loss term with a $\pi n$ periodic sine-squared function.
The model is optimized end-to-end, benefiting both detection and recognition.
We name our approach GLASS - Global to Local Attention for Scene-text Spotting.

Our method achieves state-of-the-art results on ICDAR 2015~\cite{karatzas2015icdar}, Total-Text~\cite{ch2017total}, and Rotated ICDAR 2013~\cite{liao2020spotterV3} benchmarks.
We also present blind evaluation results measured on the recently released TextOCR~\cite{singh2021textocr} dataset, largely surpassing the baseline.
GLASS is then examined across a range of text scales and orientations in an ablation study. 
Finally, we incorporate GLASS into recent E2E text spotting approaches, and show gains of 2.3\% for Mask TextSpotter v3~\cite{liao2020spotterV3} and 3.7\% for ABCNet v2~\cite{liu2021abcnetV2}, when measuring E2E F-score on Total-Text~\cite{ch2017total}.
To summarize, the  main contributions of this work are:
\begin{enumerate}
    \item A new global-to-local attention module improving text spotting performance at scale extremities
    \item A periodic orientation loss, further improving detection and recognition results across all angles
    \item State-of-the-art results on ICDAR 2015~\cite{karatzas2015icdar}, Total-Text~\cite{ch2017total}, TextOCR~\cite{singh2021textocr} and Rotated ICDAR 2013~\cite{liao2020spotterV3} benchmarks
    \item Incorporation of GLASS into other text spotting frameworks, demonstrating consistent gains
\end{enumerate}


%% file: chapters/related_work_v2.tex
\section{Background and Related Work}
\label{sec:background}

\noindent \textbf{Text Spotting}
We compare the two paradigms for text spotting, cascaded and E2E.
The cascaded option enjoys modularity, allowing to combine different architectures for detection and recognition.
By uniformly scaling and rotating the word crops to their upright orientation, the recognizer is operating on a fixed and less challenging input space.
Another benefit is that each part can train using different data.
The recognizer can leverage large amounts of synthetically generated word-crops, tailored for specific lexicons and challenging scenarios~\cite{synthtextJaderberg14c,synthadd_aaai19}, which cannot be leveraged by the detector.
For detection, synthetic images are largely limited to pre-training~\cite{gupta2016synthetic,long2020unrealtext}.
The main caveat in the cascaded approach is that no contextual information is shared between the predicted words during recognition.

In contrast, in E2E methods  the recognizer leverages contextual information from each word's surroundings, which helps disambiguate and overcome challenging scenarios. 
This is due to the large receptive field of CNN backbones~\cite{luo2016understanding}.
Furthermore, jointly training detection and recognition, benefits both tasks ~\cite{qin2019towards,wang2021towards}, leading to substantial gains.
Finally, such methods often enjoy improved latency, since the feature extraction step is done once, and shared by the detector and the recognizer.
A main drawback of E2E approaches is the limited resolution at which the recognizer operates.
The recognition branch is commonly fed with sampled features at a fixed spatial size~\cite{liao2020spotterV3,liu2021abcnetV2,baek2020crafts,qiao2020mango}, which might be insufficient for accurate prediction~\cite{lu2021master,shi2018aster,litman2020scatter}.
Specifically, the feature sampling operator, which provides the input features to the recognizer, is lossy and may fail to preserve meaningful information.
The sampling procedure is sensitive to different text scales and orientations, as discussed in ABCNet~v2~\cite{liu2021abcnetV2} and shown below.\\


\noindent \textbf{Feature Sampling}
As recognition operates on features sampled from a latent space, the  sampling procedure plays a large role in its success.
Different sampling approaches have seen several advancements over the years.

Region of Interest (RoI) Pooling~\cite{girshick2015fast} was first introduced for sampling features, and has been widely used since~\cite{li2017towards,wang2021towards}.  
It was replaced with RoIAlign~\cite{he2017mask} that used a bilinear interpolation for weighted feature sampling, that was also extended for the first time for sampling non axis-aligned (i.e., \emph{rotated}) RoIs~\cite{liu2018fots}.
For sampling arbitrarily shaped text, further extensions~\cite{lyu2018spotter,krylov2021yamts,liao2020spotterV3} added a background mask to the sampling operation for isolating the extracted word only, often relying on segmentation-based detectors or masks. 

For text, Mask TextSpotter v3~\cite{liao2020spotterV3}  
presented an anchor-free, non-parametric, segmentation proposal network where original detections are in the form of a segmentation map.
Features were sampled using hard RoI masking.
Recently, Liu \etal~introduced ABCNet~\cite{liu2020abcnet} and ABCNet v2~\cite{liu2021abcnetV2} which use a Bézier curve parametrization for localizing curved text. They exploited the parametrization using a BezierAlign operator for feature sampling. 

In the above methods, the text recognition module operates only on the limited resolution features pooled from the whole-image, the global feature map.
Our method is the first to combine additional information computed directly from a normalized word crop.
Since it is not tailored to a specific backbone or pooling layer, GLASS can be added on top of multiple existing E2E frameworks, as we show in \cref{sec:modularity}.

A few notable works predict text without relying on feature sampling.
These include CharNet~\cite{liu2018char}, which directly outputs bounding boxes of words and characters with corresponding character labels, and 
MANGO~\cite{qiao2020mango}, which divides the input image into grids and coarsely localizes the text sequence using a position-aware attention module.\\

\noindent \textbf{Global-to-Local Fusion}
There have been approaches in the literature for improving object detection performance across a large range of scales.
Recent approaches~\cite{lin2017fpn,effdet2020,hrnet2021} focused on fusing between different layers of the shared feature extraction backbone.
They harness the fact that different layers at different depths within the shared backbone have a different receptive field, and are capable of detecting details at a multitude of scales.
We leverage the global-to-local key concept from object detection~\cite{lin2017fpn,effdet2020} and adopt it for the recognition task in E2E text spotting.\\

\noindent \textbf{Orientation Prediction}
Several recent works modeled the text detection problem using a rotated box geometry for the detections.
Among the first was EAST~\cite{zhou2017east}, that suggested a hybrid approach for regressing both a rotated box and a quadrilateral around text objects.
The use of rotated RPN proposals and Rotated-ROIAlign was first suggested in~\cite{ma2018arbitrary}, using the regular $L_1$ loss to regress the output boxes. \amir{boxes vs. proposals will confuse non-detection people here}
The authors in~\cite{zhu2019rotated}, identified an ambiguity in the angle prediction, namely that the same box can be described by four valid angles, a different angle perpendicular for each face. 
They tackled it with a cascaded process where a single correct box orientation is regressed in a gradual manner. 
In~\cite{lee2021rotated} the same ambiguity was handled by optimizing over the minimal angle difference among all of the detected box sides, and in~\cite{qian2021learning} the ambiguity is dealt with by representing the orientation of each box with 8 parameters, and regressing over all of them, while ensuring continuity of the loss function.
Our approach tackles the angle regression ambiguity by introducing an orientation-aware, periodic trigonometric loss, as further discussed in \cref{sec:detection}.


%% file: chapters/method_v2.tex
\section{Method}
\label{sec:method}
\begin{figure}[t]
 \centering
  \includegraphics[width=0.89\linewidth]{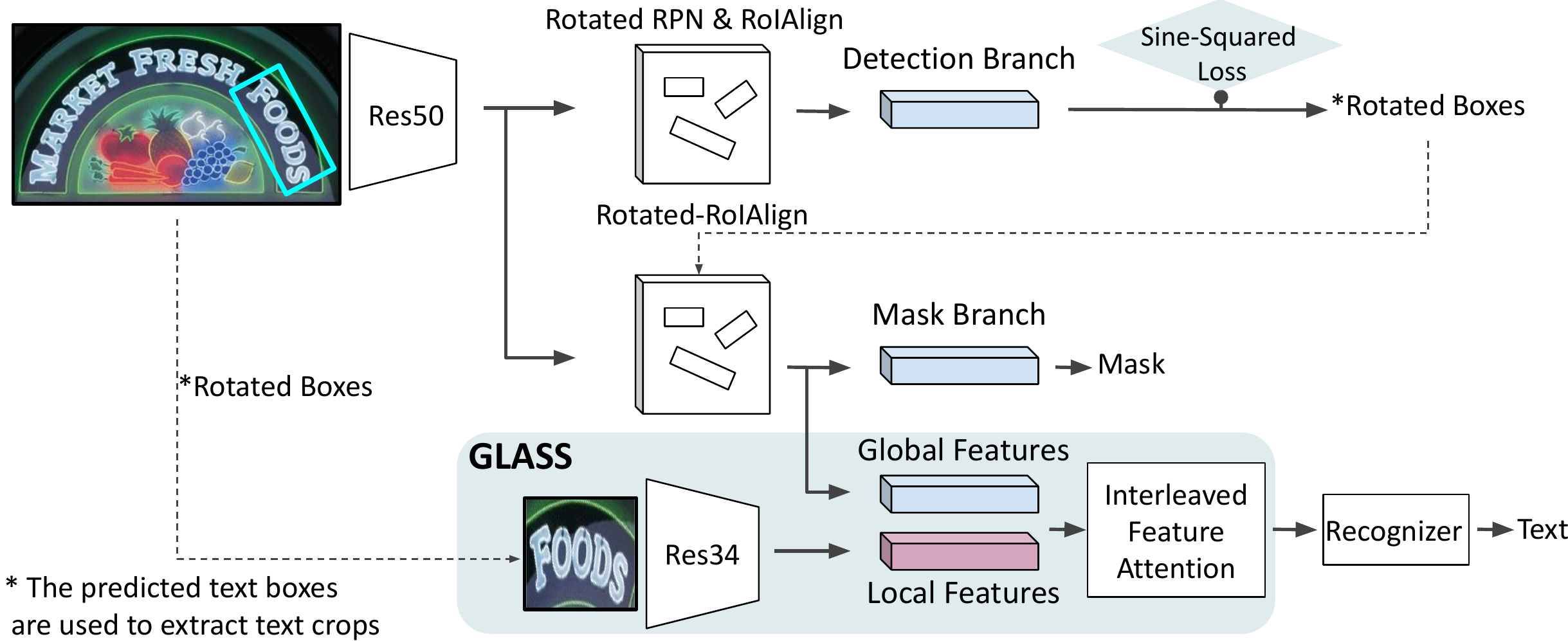}
  \caption{\textbf{Global to Local Scene-text Spotting.} The global detection branch is a Mask R-CNN variant supervised by our novel sine-squared loss for rotated box prediction. During inference, predicted boxes are used to sample global backbone features and to crop the original image as input for the local feature extractor. Global and local features are fused in GLASS and passed to the recognizer for transcript extraction.}
  \label{fig:framework}
\end{figure}
\subsection{GLASS Fusion Module}
\label{sec:GLASS}
Our pipeline is composed of three principal components, seen in \cref{fig:framework}: the detection branch, the GLASS fusion module, and the recognition head.
The detection branch is used for locating words,  predicting their bounding boxes and segmentation masks.
It is trained with the added orientation-aware loss, and its backbone is used for extracting the global features.
The fusion module combines the global and local features, producing an enriched embedding that is then fed into the recognition head.
In this work, we use a Rotated Mask R-CNN~\cite{he2017mask,ma2018arbitrary} as the baseline approach for our detection branch, and for the recognition branch we use ASTER~\cite{shi2018aster}.

We begin by presenting our global-to-local fusion module in \cref{sec:GLASS}, 
and follow with a description of our orientation-aware loss in \cref{sec:detection}.
Finally, we discuss aspects regarding the overall architecture and training objective in \cref{sec:e2e}.

We propose a fusion module for incorporating the scale and rotation invariance of the local word-crop approach into an end-to-end text spotter, while still using global context.
Uniformly scaled and aligned crops abstract away nuisances such as the original word's size and off-axis rotation, and allow us to maximize our ability to extract text.

The predicted boxes of the detection branch are used in two sampling operations.
For \emph{global} features, we sample the FPN~\cite{lin2017feature} features from the detection branch.
For \emph{local} features, we sample the image directly (i.e. crop), performing an affine transformation that yields a uniformly scaled, axis-aligned word crop.
This crop is passed through a local feature extractor. 

\begin{figure}[t]
 \centering
  \includegraphics[width=1.0\linewidth]{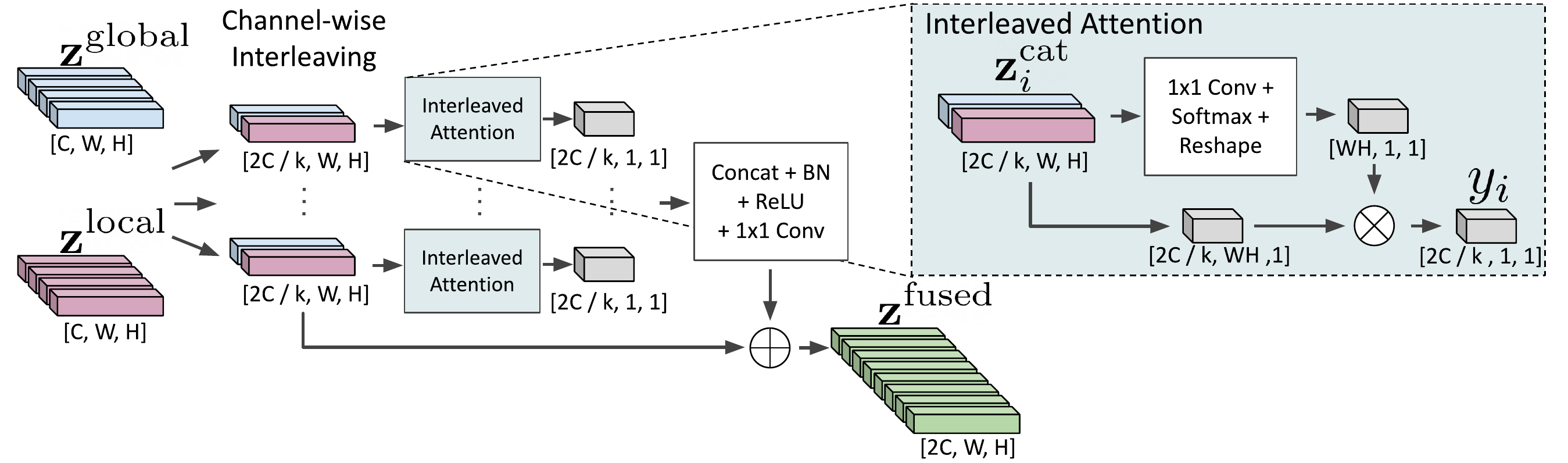}
  \caption{\textbf{Global to Local attention feature fusion.} GLASS takes both \emph{global} (image) and \emph{local} (crop-level) features as input, and outputs the fused feature map.
  The input feature maps are channel-wise interleaved, concatenated and split into $k$ blocks. Each block is processed by an attention module, producing $k$ tensor outputs. These are then concatenated, transformed by $1\times 1$ convolution and summed element-wise with the input feature maps.
  }
  \label{fig:featrefusion}
\end{figure}

Formally, we denote the input image $\mathbf{x}$ and its FPN features $\mathbf{z}$.
Following detection, the global feature map $\mathbf{z}$ is sampled using the predicted boxes, yielding $\mathbf{z}^{\rm global}$.
Using the same boxes, the image $\mathbf{x}$ is cropped and aligned into $\mathbf{x}^{\rm local}$, which in turn is embedded using a shallow dedicated backbone into $\mathbf{z}^{\rm local}$.
Every text detection is now represented by two separate feature maps, $\mathbf{z}^{\rm global}$ and  $\mathbf{z}^{\rm local}$,
illustrated in~\cref{fig:featrefusion} by light-blue and pink bars, respectively.

Inspired by~\cite{lu2021master}, we propose an interleaved attention procedure that operates over small feature blocks, aiming to combine and use the most relevant information for the text recognition task, from both input features.
Attending over small blocks is significantly lighter than standard attention mechanisms in high dimensionality, and is shown to improve robustness for downstream tasks~\cite{lu2021master}.
The interleaved attention combines global and local features in a learned way, maximizing informational content.
This dynamic weighting allows the attention mechanism to place greater emphasis on specific relevant context, depending on the input.

The global and local features are first combined to $k$ block tensors by an interleaved concatenation, where $k 
\ll C$.
The $i$th block is given by
\begin{align}
\mathbf{z}^{\rm cat}_i &= \left[ z^{\rm global}_{i\cdot m+1},\ z^{\rm local}_{i\cdot m+1},\ \dots,\ {z}^{\rm global}_{i\cdot m + m},\ {z}^{\rm local}_{i\cdot m + m}\right],
\end{align}
where $m=\lceil C/k \rceil$.
Block indices are given by $i \in \{0, 1, \dots, k-1\}$, and ${z}^{\rm global}_j,{z}^{\rm local}_j$ depict the $j$th channel of $\mathbf{z}^{\rm global},\mathbf{z}^{\rm local}$, for $j\in[1,C]$.

A spatial attention operator is then applied to each of the $k$ blocks in $\mathbf{z}^{\rm cat}$, as shown in \cref{fig:featrefusion} within the dashed box, such that
\begin{align}
\label{eq:attention}
y_i = {\rm vec}(\mathbf{z}^{\rm cat}_i)^T
{\rm vec}\left( \operatorname{softmax}(v_i^T \mathbf{z}^{\rm cat}_i) \right), 
\end{align}
yielding an attentional vector $y_i\in {\mathbb R}^{2C/k}$.
Here ${v}_i \in {\mathbb R}^{2C/k}$ is a learnable vector and ${\rm vec}(\cdot)$ reshapes a tensor of size $(C,W,H)$ into a matrix of size $(C,WH)$.
Interleaving the two feature maps ensures that in 
\cref{eq:attention} we mix information that corresponds with both global and local features.

Next we stack the $k$ attention vectors $y_{1...k}$ channel-wise, apply batch normalization (BN), ReLU and a $1\times 1$ convolution for capturing channel-wise dependencies, resulting in the tensor $\mathbf{y}$. 
The fused output is an element-wise addition of $\mathbf{y}$ and the interleaved-concatenated feature maps, $\mathbf{z}^{\rm cat}$,
illustrated by the green bars in \cref{fig:featrefusion}. Formally, 
\begin{align}
    \mathbf{z}^{\rm fused} = \mathbf{z}^{\rm cat} + \mathbf{w}^T\operatorname{ReLU}(\operatorname{BN}(\mathbf{y}))\,,
\end{align}
where $\mathbf{w}\in{\mathbb R}^{C\times H\times W}$ are learnable weights.
The output $ \mathbf{z}^{\rm fused}$ is then used as input to the recognition head.

We note that there are two alternatives to the proposed interleaved attention.
The first is a na\"ive concatenation of the global and local features, which is tested in \cref{sec:detection}, and is shown to reduce accuracy.
The second is performing a full attention computation between the full dimension of local and global features, however, this computation is unfeasible due to computational limitations.

\subsection{Orientation Prediction}
\label{sec:detection}

Unlike objects in other common object-detection benchmarks such as COCO~\cite{mscoco2014} and Pascal VOC~\cite{pascalvoc}, text instances are long, narrow and directed.
A word extracted upside-down or rotated by $90^{\circ}$ is usually non-recoverable in terms of recognition.
This makes orientation prediction especially important and meaningful.
To this end, we propose a new orientation-aware loss function operating on rotated box detections $\mathbf{r}\in\mathbb{R}^{N\times 5}$, where the first 4 coordinates describe the rotated box center, width and height, and the last coordinate, $\theta\in\mathbb R$, depicts its upward facing angle.
The loss function for the $m$th matched detection is given by
\begin{align}
    \mathcal{L}_\text{rbox} & = \sum_{i=1}^{4} \alpha_i \left| \hat{r}_i - r_i \right| + \alpha_\theta \sin^2 \left( \hat{\theta} - \theta \right),
\label{eq:rbox_loss}
\end{align}
where the hat denotes prediction.
The constants $\alpha_i$ for $i\in[1,4]$ and $\alpha_\theta$ are chosen empirically.
The sine-squared function has a periodicity of $n\pi$ for $n \in \mathbb{Z}$, leveraging the fact that a rotated rectangle is symmetric to $n\pi$ rotations.
This symmetry removes an inherent ambiguity during training, allowing the same prediction for boxes that are either upright or flipped upside-down.
This mechanism was empirically shown to improve detection results across all angular range, as further explored in \cref{sec:angle_analysis} and shown in \cref{fig:angle_analysis}.
For each word, the orientation angle is then used to perform a rotated pooling operation on the shared backbone features, yielding the global feature input.
This process is common in additional E2E frameworks~\cite{liu2021abcnetV2,liao2020spotterV3}, but without the orientation-aware loss, orientation mistakes in the form of discrete jumps of $k\pi/2, k\in\mathbb{Z}$ degrees are more common.
In our implementation, the predicted angle is also used to generate an  oriented word-crop, from which the local features are computed, as illustrated in \cref{fig:framework}.

\subsection{Global to Local End-to-end Text Spotting}
\label{sec:e2e}
Here, we describe our proposed E2E framework 
with GLASS, shown in \cref{fig:framework}.
For the shared backbone, we use the commonly used ResNet50 and FPN~\cite{lin2017feature}.
Its associated features $\mathbf{z}^{\rm global}$ are sampled using Rotated-RoIAlign
operating directly on the FPN levels as in \cite{qin2019towards}.
For obtaining the local feature maps $\mathbf{z}^{\rm local}$, we first sample a crop of the text RoI from the input image using Rotated-RoIAlign layer.
Then, the crop features are extracted by  ResNet34 backbone~\cite{focusing_att2017}.

Finally, we fuse the global and local feature maps using the interleaved attention operation, described in \cref{sec:GLASS}, yielding $\mathbf{z}^{\rm fused}$, which is the recognition module's input.
The recognition head, detailed in the Supplementary Material, provides the transcript for each word.
We note that the mask head is used as a parallel branch to the recognizer, and only receives $\mathbf{z}^{\rm global}$ as input.

The overall loss function $\mathcal{L}$ used for the E2E supervised training is given by
\begin{equation}
    \mathcal{L} =  \mathcal{L}_{\rm rbox} +  \lambda_1  \mathcal{L}_{\rm mask} + \lambda_2  \mathcal{L}_{\rm rec}
    \;.
   \label{eq:total_loss}
\end{equation}
Here, the mask loss $\mathcal{L}_{\rm mask}$ is identical to Mask R-CNN~\cite{he2017mask} and $\mathcal{L}_{\rm rec}$ is the recognition loss.

We note that GLASS has no effect on the computational cost of the detector, including its mask branch, and the recognizer head.
The computational aspects of GLASS, as well as the loss terms used in training, are further discussed in the Supplementary Material.


%% file: chapters/experiments_v2.tex
\section{Experiments}
\label{sec:experiments}

We evaluate the performance of our method on several benchmarks, testing our method's robustness to rotations and text size.
First, we compare our full framework with GLASS to current art.
Next, we examine GLASS when integrated into two common E2E text spotting architectures, Mask~TextSpotter~v3~\cite{liao2020spotterV3} and ABCNet~v2~\cite{liu2021abcnetV2}.
Finally, we provide a comprehensive ablation study isolating the contribution of GLASS for different data distributions with various settings.
Additional ablation studies are presented in the Supplementary Material.

\subsection{Datasets}
\label{sec:datasets}
\textbf{SynthText}~\cite{gupta2016synthetic} is a synthetically generated dataset containing approximately 800K images and 6M synthetic text instances.
\textbf{ICDAR 2013}~\cite{karatzas2013icdar} has 233 testing images containing mostly horizontal text. We synthetically rotate these images by various angles and use it to measure our performance on rotated text.
\textbf{ICDAR 2015}~\cite{karatzas2015icdar} consists of 1,000 training images and 500 testing images.
Most of the images are of low resolution and contain small text instances.
\textbf{Total-Text}~\cite{ch2017total} contains 1,255 training and 300 testing images.
It offers text instances in a variety of shapes, including horizontal, rotated, and curved text. 
\textbf{TextOCR}~\cite{singh2021textocr} is a recently published arbitrary-shaped detection and recognition dataset containing of 21,778 train, 3153 validation and 3232 test images with more than 700k, 100k and 80k annotated words, respectively. 

\subsection{Implementation details}
\label{sec:implementation}
We follow the common SynthText pre-training scheme~\cite{liao2020spotterV3,baek2020crafts}.
For Total-Text, we fine-tune using a mixture of Total-Text and SynthText datasets, as in~\cite{baek2020crafts}.
For ICDAR13 and ICDAR15 results, we train also on both datasets, following~\cite{liao2020spotterV3}.
For TextOCR results, we follow the baseline~\cite{singh2021textocr} and use all of the datasets mentioned in \cref{sec:datasets}. 
In the ablation studies (\cref{sec:ablation_study}), we fine-tune the model for 100k iterations with a batch size of 8 images.
In \cref{sec:results}, the model is fine-tuned for 250k iterations with a batch size of 24 images.
The recognizer used is an off-the-shelf component, based on ASTER~\cite{shi2018aster}.
Additional implementation details are found in the Supplementary Material.

\subsection{Comparison with State-of-the-Art}
\label{sec:results}
\input{tables/unified}
Quantitative results for end-to-end text recognition on the ICDAR15, Total-Text and TextOCR datasets are listed in \cref{tab:res_icdar_totaltext}.
For ICDAR15, our method outperforms previously reported word spotting protocol results for all three lexicons, and for the end-to-end evaluation protocol with Generic lexicon. 
For the Total-Text dataset, our method achieves state-of-the-art F-measure results for both settings in the word spotting evaluation, and for full-lexicon end-to-end.
For no-lexicon end-to-end, GLASS outperforms all methods but CRAFTS~\cite{baek2020crafts}.

We are among the first to report results for the challenging TextOCR test dataset in \cref{tab:res_icdar_totaltext}. This newly released dataset is an order of magnitude larger than previous ones and has ample variation in scale and rotation.
Thresholds for the detection and recognition heads were set using the TextOCR validation set.  
Our method achieves state-of-the-art F-measure results, surpassing Mask TextSpotter v3~\cite{liao2020spotterV3} by 16.3\% on end-to-end evaluation protocol.
Both methods were optimized with a similar data profile, including TextOCR train data~\cite{singh2021textocr}.

\begin{figure}[t]
 \centering
  \includegraphics[width=1.0\linewidth]{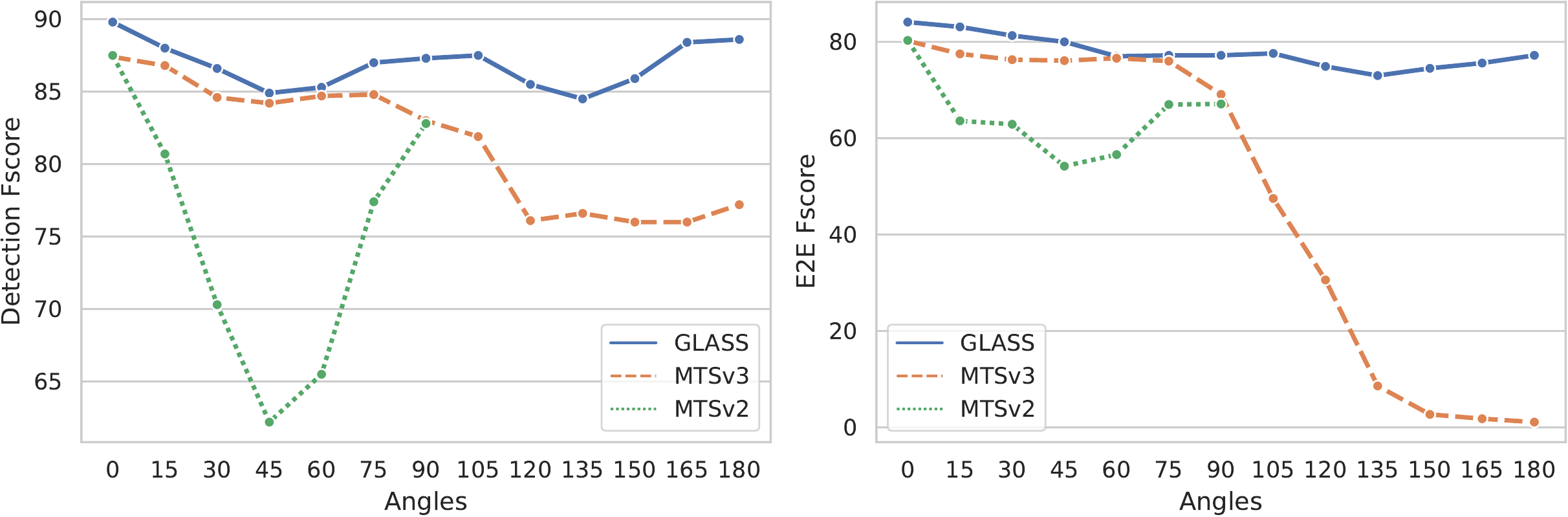}
   \caption{\textbf{GLASS contribution for different angles.}
   We measure the performance on the Rotated ICDAR 2013 dataset.
   Combining GLASS with our novel sine-squared loss improves text detection and recognition across all angles.
   Notice the $4\%$ detection and $7\%$ recognition gains at angles close to $90^{\circ}$. 
   }
  \label{fig:angle_analysis}
\end{figure}

To validate our framework performance on oriented text, we show results on the Rotated ICDAR 2013 benchmark~\cite{liao2020spotterV3} in \cref{fig:angle_analysis}.
Our approach with GLASS and the sine-squared loss outperforms previous art on both text detection and recognition across all angles, and especially benefits the detection algorithm on steep angles larger than $60^{\circ}$.  

\subsection{Incorporating GLASS into other methods}
\label{sec:modularity}
GLASS can be incorporated into any text spotting architecture that uses a feature pooling module, regardless of its specific pooling mechanism.
To demonstrate this, we employ GLASS in two common E2E text spotting works, Mask TextSpotter v3~\cite{liao2020spotterV3} and ABCNet v2~\cite{liu2021abcnetV2}.
We note that Mask TextSpotter v3 and ABCNet v2 use different backbones and RPN modules, different detection and recognition heads, in addition to other minor differences.
Importantly, unlike our method which uses Rotated-RoIAlign pooling,  Mask TextSpotter v3 uses an axis-aligned RoIAlign with hard feature masking, and ABCNet v2 applies BezierAlign.
Despite the differences between all three architectures, the use of GLASS within 
both Mask TextSpotter v3 and ABCNet v2 is straightforward and requires minimal changes. 

For each method, we use its publicly available source code\footnote{https://github.com/MhLiao/MaskTextSpotterV3}\footnote{https://github.com/aim-uofa/AdelaiDet}, and train both architectures with and without the GLASS component, following the training procedure published by the respective authors.
Results are shown in \cref{tab:modularity}.
Adding only minor computational overhead, GLASS provides a considerable benefit to the E2E performance of each method.
Further discussion on the computational aspects of GLASS is found in the Supplementary Material.

\input{tables/modularity}

\subsection{Ablation study}
\label{sec:ablation_study}

To evaluate the effectiveness of individual parts in our proposed framework, we conduct ablation studies on the Total-Text and ICDAR15 datasets.
In \cref{tab:ablation1,tab:ablation3,tab:scale_analysis1,tab:ablation2} we report the \emph{end-to-end} F-measure (Hmean) as defined in ICDAR15~\cite{karatzas2015icdar}.
Every model version is pre-trained and fine-tuned as an end-to-end system independently for every experiment.

The \textbf{baseline} architecture consists of a Mask R-CNN  detection branch  with a Rotated-RoIAlign component and our novel rotated box regression loss, as described in \cref{sec:detection}.
The recognizer  is set as described in Supplementary Material and receives as input only the \emph{global} features from the shared backbone, as described in~\cref{sec:GLASS} and shown in \cref{fig:main}b.
The recognition head remains unchanged for all experiments, with the sole difference being the input features selected for it.\\

\noindent \textbf{Contribution of GLASS to end-to-end performance}
The effect of GLASS on overall performance is presented in \cref{tab:ablation1}.
Different feature map and fusion combinations are compared, all using the baseline architecture for detection.
Replacing the global features with local features as recognition input, shown in the second row of the table, causes a 3.2\% and 4.6\% regression for Total-Text and ICDAR15 datasets, respectively.
We identify the main reason for this regression in a noticeable drop in the detection performance (see in Supplementary Material), and the lack of mutual supervision of the detection-recognition.

A simple channel-wise concatenation of global and local features (row~3) improves the results over the baseline by 2.4\% and 3.5\% on Total-Text and ICDAR15 datasets.
Furthermore, using GLASS provides further boosts for both Total-Text and ICDAR15 datasets, reaching 3.1\% and 4.6\% over the baseline.
Overall, adding the GLASS module leads to considerable gains in the E2E performance, while only 
increasing latency by roughly $10\%$.\\
\input{tables/ablation1.tex}
\input{tables/ablation3.tex}
\noindent \textbf{Orientation robustness analysis}
\label{sec:angle_analysis}
We compare our framework with two different orientation losses: our proposed sine-squared loss in \cref{eq:rbox_loss} and the commonly used $L_1$ loss. Both models use GLASS.
The results on Total-Text and ICDAR15 datasets are presented in \cref{tab:ablation3}.
Relying on the $L_1$ loss, instead of our  $\sin^2$  loss, leads to a result regression of 0.7\% on the Total-Text dataset, which emphasizes arbitrarily rotated text.

\input{tables/total_text_scale}

\input{tables/ablation2.tex}

\begin{wrapfigure}[12]{R}{0.25\textwidth}
\centering
  \includegraphics[width=1.0\linewidth]{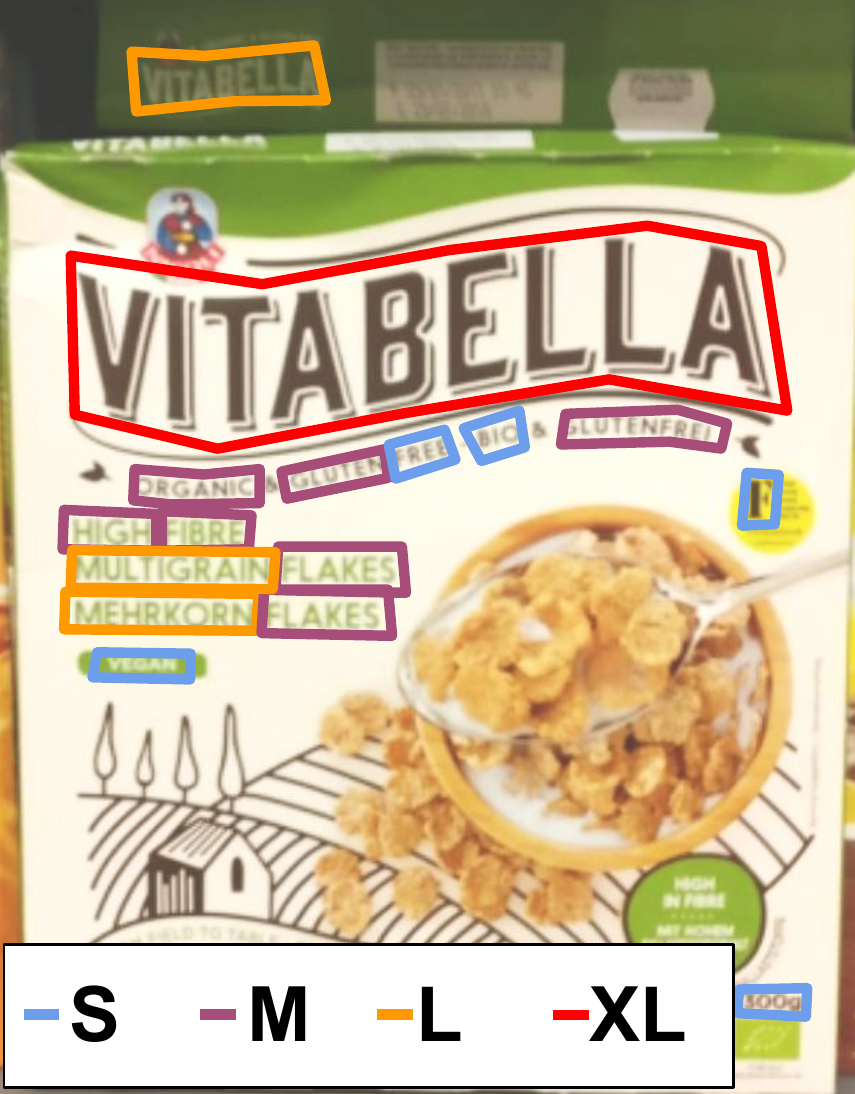}
  \caption{Illustration of text scale groups.  
}
\label{fig:scale_illustrations}
\end{wrapfigure}

\noindent \textbf{Scale robustness analysis}
We establish our method's contribution in challenging cases in both quantitative and qualitative manners.
The relative contribution of GLASS at different scales is quantified by performing a custom ablation study using the Total-Text dataset. 
Total-Text contains a variety of challenging texts at different scales and rotations.
First, we divide text instances into four groups: small, medium, large and extra-large, denoted by S, M, L and XL accordingly, and illustrated in \cref{fig:scale_illustrations}.
The groups are defined as four equally-sized bins w.r.t. the square root area of their ground-truth text polygon, over the entire dataset.

The F-score is measured for each population of prediction and ground-truth polygons independently,  shown in \cref{tab:scale_analysis1}. 
As expected, most of the gain is achieved over the small and extra-large text groups, compared to the baseline that predicts text solely using the global branch.

\begin{figure*}[t]
 \centering
  \includegraphics[width=1.0\linewidth]{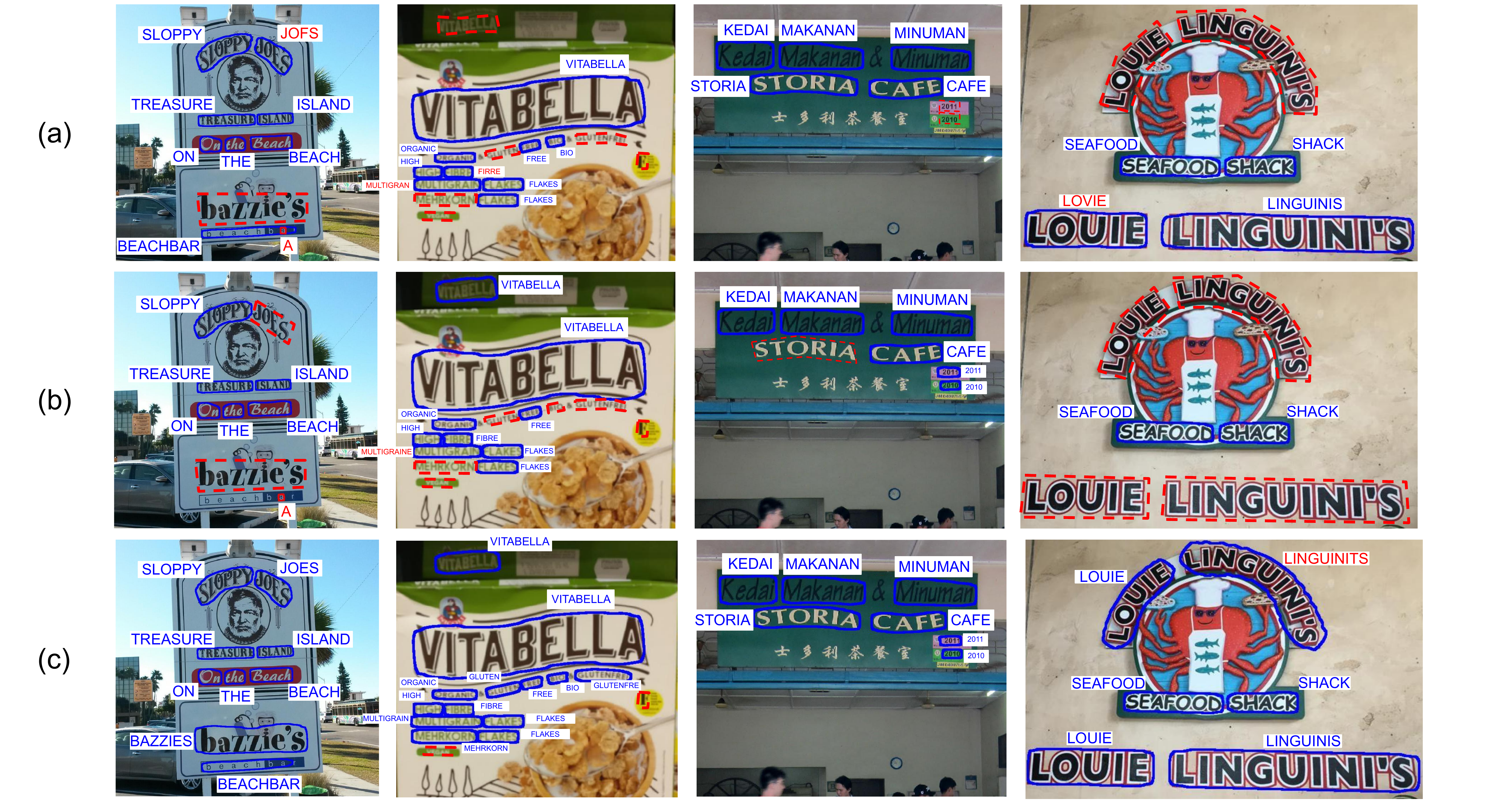}
   \caption{\textbf{Qualitative results for Total-Text.}
  Predictions from: \textbf{(a)}  A standard E2E text spotting framework. \textbf{(b)} An E2E framework using only the local features for recognition, and \textbf{(c)} Our proposed method with the GLASS component. Blue and red represent correct and incorrect predictions, respectively. GLASS improves recognition, specifically for small and large words, matching the results in \cref{tab:scale_analysis1}.} 
  \label{fig:totaltext_qualitative}
\end{figure*}

\setlength\intextsep{0pt}
A qualitative comparison is shown in \cref{fig:totaltext_qualitative}, where we picked examples from the Total-Text dataset that contain a mix of small and large scale text.
The baseline result, shown in row (a), is under-performing on challenging text instances that are either small, very large or have a steep rotation angle.
In row (b), where only the local features are used, there is a notable regression in both detection and recognition accuracy.
In row (c) our method is robust to both scale and orientation, and capable of accurately detecting and recognizing text in extreme and challenging scenarios.\\

\noindent \textbf{Isolating recognition branch performance}
Lastly, we assess the impact of the GLASS module on the recognition task by injecting the ground-truth rotated boxes as the detection output.
By overriding the entire detection branch with \emph{oracle} predictions we are able to isolate the recognizer and compare multiple configurations.
The results are presented in \cref{tab:ablation2}, showing that using the fused features from GLASS contributes to a large increase of $5.2\%$ and $7.2\%$ in the recognition performance on Total-Text and ICDAR15 respectively.

%% file: tables/unified.tex
\begin{table}[t]
\caption{\textbf{Results for ICDAR 2015, Total-Text and TextOCR datasets.} ‘S’, ‘W’ and ‘G’ refer to strong, weak and
generic lexicons. “None” refers to recognition without any lexicon. “Full” lexicon contains all the words in the test set. (*) refers to using specific lexicons from~\cite{liao2020spotterV3}. 
($^\dagger$)~indicates IoU of 0.1 was used instead of 0.5 during evaluation.
($^\ddagger$)~represents results obtained using method's official source code.}
\centering
\resizebox{\textwidth}{!}{
\begin{tabular}{lccccccccccc}
\toprule

\multirow{3}{*}{Method}   & \multicolumn{6}{c}{ICDAR 2015} &  \multicolumn{4}{c}{Total-Text} &  \multicolumn{1}{c}{TextOCR}  \\
 & \multicolumn{3}{c}{Word Spotting} & \multicolumn{3}{c}{End-to-End} &  \multicolumn{2}{c}{Word Spotting} & \multicolumn{2}{c}{End-to-End} & \multirow{2}{*}{End-to-End} \\ 
 
                  &        S         & W         & G         & S            & W           & G  &      None & Full  & None & Full  &   \\ \hline
   
 TextDragon~\cite{feng2019textdragon}    &     86.2      &    81.6       &     68.0       &         82.5         &   78.3   & 65.2   & - & -        & 48.8                & 71.8   & -      \\
ABCNet v2~\cite{liu2021abcnetV2}    &    -      &    -       &     -       &         82.7         &   78.5   & 73.0   &   70.4                   &    78.1       & - & -    & -      \\
MTSv3*~\cite{liao2020spotterV3}     & 83.1      & 79.1      & 75.1       & 83.3         & 78.1        & 74.2        & - & -          & 71.2                 & {78.4}     & 50.8             \\
Text Perc.~\cite{qiao2020text}           & 84.1        & 79.4        & 67.9  & 80.5      & 76.6     & 65.1                  & 69.7                 & {78.3}        & - & -   & -             \\

CRAFTS~\cite{baek2020crafts}        & -         & -         & -          & 83.1         & \textbf{82.1}        &     \underline{74.9}    &- & -   &\textbf{78.7}        & -            & -            \\
MANGO*$^\dagger$~\cite{qiao2020mango}        & {85.2}      & 81.1      & 74.6      & \textbf{85.4}         & \underline{80.1}        & 73.9                         & \underline{72.9}                 & \underline{83.6}     & 68.9$^\ddagger$	 & \underline{78.9}$^\ddagger$ & -    \\
YAMTS*~\cite{krylov2021yamts}      & \textbf{86.8}      & \underline{82.4}      & \underline{76.7}       & \underline{85.3}         & 79.8        & 74.0               & - & -                         & 71.1                 & {78.4}      & -    \\  \hline

\multicolumn{1}{l}{\textbf{Ours}*} 
&   \textbf{86.8}
&      \textbf{ 82.5 }
&    \textbf{78.8}
&    {84.7}
&     \underline{80.1}
&    \textbf{76.3} 
         & \textbf{79.9} & \textbf{86.2} &  \underline{76.6}
&   \textbf{83.0} & \textbf{67.1}
 \\
\bottomrule
\end{tabular}}
\label{tab:res_icdar_totaltext}
\end{table}

%% file: tables/modularity.tex
\begin{table}[t]
\centering
\caption{\textbf{GLASS results with Mask TextSpotter v3 (MTSv3)}~\cite{liao2020spotterV3} \textbf{and ABCNet v2}~\cite{liu2021abcnetV2}\textbf{.} First and third rows are results reproduced using the official MTSv3 and ABCNet v2 implementations. The second and fourth rows show the effect of incorporating GLASS into MTSv3 and ABCNet v2. }
\label{tab:modularity}

\begin{tabular}{l cc cc} 
\toprule

\multicolumn{1}{c  }{\multirow{2.5}{*}{Method}} & \multicolumn{2}{c}{{Total-Text}} & \multicolumn{2}{c}{{ICDAR 2015}} \\
\cmidrule(lr){2-3} \cmidrule(lr){4-5}  
 & {E2E Hmean} & { FPS} & {E2E Hmean} & { FPS} \\
\midrule	

MTSv3~\cite{liao2020spotterV3}  & 67.5 & 2.2  & 68.5 & 2.6 \\
   ~~~~ + GLASS                        & 69.8 \textbf{(+2.3)} & 2.0  & 72.3 \textbf{(+3.8)} & 2.3 \\ 
\midrule
ABCNet v2~\cite{liu2021abcnetV2} & 67.6 & 6.5  & - & - \\
  ~~~~ + GLASS                         & 71.3 \textbf{(+3.7)}  & 6.0  & - & - \\
\bottomrule
\end{tabular}
\end{table}


%% file: tables/ablation1.tex
\begin{table}[t]
\centering
\caption{\textbf{Ablation study - Fusion.} ``Global'' and ``Local'' columns stands for the use of image-level (Global) and cropped (Local) features.
``Fusion'' column compares two different fusion operations where both features are used: a simple channel-wise concatenation and our fusion method. All rows use the Rotated Mask R-CNN detector and recognition head described in \cref{sec:detection} and \cref{sec:e2e}. ``FPS'' column states the average latency in frames-per-second measured for Total-Text. We measure End-to-End Hmean on Total-Text (TT) and ICDAR15 (IC15).}
\label{tab:ablation1}
{
\begin{tabular}{c c c c c c c}
\toprule

%
&  Global & Local & \multicolumn{1}{c}{Fusion} & \multicolumn{1}{c}{TT} & \multicolumn{1}{c}{IC15} & FPS \\
\hline
Baseline  & \checkmark &   &    --    & 72.6 & 69.1 & \multicolumn{1}{c}{3.0} \\ 
Baseline + Local  &   & \checkmark      &  --   & 69.4  ($\downarrow 3.2$) &  64.5 ($\downarrow 4.6$) & \multicolumn{1}{c}{2.8} \\ 
Baseline + Global-Local & \checkmark & \checkmark & Concat. & 75.0 ($\uparrow 2.4$) &  72.6 ($\uparrow 3.5$) & \multicolumn{1}{c}{2.7} \\ 
\textbf{Baseline + GLASS}&  \checkmark & \checkmark &     Ours & 75.7 ($\uparrow 3.1$)& 73.7 ($\uparrow 4.6$) & \multicolumn{1}{c}{2.7} \\ \bottomrule
\end{tabular}}
\end{table}

%% file: tables/ablation3.tex
\begin{table}[t]
\centering
\caption{\textbf{Ablation study - Orientation loss.}
We use Total-Text and ICDAR15 to compare the two losses for the rotated box angle: (a) the commonly used $L_1$ loss and (b) our novel sine-squared loss from \cref{eq:rbox_loss}.
Both experiments use GLASS. The metrics R, P, and H denote detection recall, precision, and Hmean respecively, while E2E denoted End-to-End Hmean.}
\label{tab:ablation3}
\begin{tabular}{ccccccccc}
\toprule
\multicolumn{1}{c}{\multirow{3}{*}{}} & \multicolumn{4}{c}{\underline{Total-Text}}& 
\multicolumn{4}{c}{\underline{ICDAR 2015}} \\
\multicolumn{1}{c}{}                        & R      & P      & H      &   E2E       & R      & P      & H      &         E2E      \\ \hline

GLASS + {$L_1$ loss}      & \textbf{86.3}   &  88.2   &  87.2             &  75.0  & 83.4   &  85.0   &  84.2             &  73.5
\\
GLASS +  {$\sin^2$ loss}      & 85.5  & \textbf{90.8} & \textbf{88.1}              & \textbf{75.7}   & \textbf{84.5}   &  \textbf{86.9}   &  \textbf{85.7}             &  \textbf{73.7} 
\\ \bottomrule
\end{tabular}

\end{table}

%% file: tables/total_text_scale.tex
\begin{table}[t]
\centering
\caption{\textbf{GLASS contribution across different scales.}
  We measure the performance over 4 scale groups in Total-Text: small, medium, large and
extra-large, denoted by S, M, L, XL accordingly
(see \cref{fig:scale_illustrations}).
    The specific sizes were chosen for creating equally sized bins of ground-truth instances.
  GLASS outperforms the other configurations, especially on the lower and higher end of the scales. The baseline achieves comparable performance to GLASS on medium and large scales.}
\begin{tabular}[b]{ccccc}
\toprule
\multicolumn{1}{c}{}  & 
\multicolumn{1}{c}{S}  &
\multicolumn{1}{c}{M}  &
\multicolumn{1}{c}{L}   &
\multicolumn{1}{c}{XL}  
\\ 
\hline
 \multicolumn{1}{c}{Baseline}&   71.8 & 77.8 & 77.7 & 79.0 \\
 \multicolumn{1}{c}{Baseline + Local branch}& {\!\! 72.7 ($\uparrow 1.9$) \!\!} & {\!\! 77.3 ($\downarrow 0.5$) \!\!} & {\!\! 74.9 ($\downarrow 2.8$) \!\!} & {\!\! 75.5 ($\downarrow 3.5$) \!\!} \\
 \multicolumn{1}{c}{\textbf{Baseline + GLASS}}& {\!\! \textbf{73.9} ($\uparrow 2.1$) \!\!} & {\!\! \textbf{78.1} ($\uparrow 0.3$) \!\!} & {\!\! \textbf{77.8} ($\uparrow 0.1$) \!\!} & {\!\! \textbf{80.8}  ($\uparrow 1.8$) \!\!} 
\\ \bottomrule
\end{tabular}
\label{tab:scale_analysis1}
\end{table}

%% file: tables/ablation2.tex
\begin{table}[h]
\centering
\caption{\textbf{Ablation study - Recognition with ground-truth boxes.} To isolate the recognizer's performance, \emph{ground-truth} boxes are used, simulating perfect detections. We compare global, local and fused recognizer inputs on Total-Text and ICDAR15.}
\label{tab:ablation2}
\begin{tabular}{c c cc c c}
\toprule
\multicolumn{1}{c}{} & \multicolumn{1}{c}{GT} & \multicolumn{2}{c}{Features} &  \multicolumn{1}{c}{ Total-Text} & \multicolumn{1}{c}{ICDAR15}\\ 
\multicolumn{1}{c}{} & \multicolumn{1}{c}{Detection}& \multicolumn{1}{c}{ Global} & \multicolumn{1}{c}{Local}                  & Hmean           & Hmean        \\
\hline
\multicolumn{1}{c}{Baseline}& \checkmark& \checkmark  &   \multicolumn{1}{c}{}  & \multicolumn{1}{c}{75.3}  & \multicolumn{1}{c}{72.8}   \\
 \multicolumn{1}{c}{Baseline + Local branch}& \checkmark&    &    \multicolumn{1}{c}{\checkmark }   & \multicolumn{1}{c}{\!\! 74.2 ($\downarrow 1.1$) \!\!}      & \multicolumn{1}{c}{\!\! 78.5 ($\uparrow 5.7$) \!\!}   \\
 \multicolumn{1}{c}{\textbf{Baseline + GLASS}}& \checkmark&  \checkmark  &  \multicolumn{1}{c}{\checkmark  }   & \multicolumn{1}{c}{80.5 ($\uparrow 5.2$)}  & \multicolumn{1}{c}{80.0 ($\uparrow 7.2$)} \\
 \bottomrule
\end{tabular}
\end{table}

%% file: chapters/discussion.tex
\section{Discussion}
\label{sec:discussion}

We propose two extensions for existing text spotting methods. First is combining global and local features for end-to-end text recognition, and a fusion operator enabling that, termed GLASS.
The other is an orientation prediction side task, using the orientation-aware sine-squared objective during optimization.

The proposed algorithm combines highly-contextual global features, which also encode each word's surroundings and allows reading it in-context, like humans do, with uniformly scaled and oriented local features, abstracting away scale and rotation. This improves performance for common cases, and even more so for cases of strong scale and rotation.

Extensive experiments over four benchmarks, including the challenging Rotated ICDAR 2013 and the new TextOCR show state-of-the-art results.
Ablation studies highlight our contribution over scale and rotation ranges, as well as our method's applicability to other recent text spotting methods.

%% file: supplementary/supp_content.tex

In the supplementary material, we share further details and analysis for our work.
In~\cref{sec:computational}, we provide estimates for the latency and computations costs added when using GLASS.
In~\cref{sec:technical_details}, we discuss our recognition head selection and further implementation details of our main method.
Finally, in~\cref{sec:further}, we explore GLASS's effect over detection performance, share results for the TextOCR validation set, present failure cases and further explore GLASS's contribution across different scales.


\section{GLASS Computational Cost}
\label{sec:computational}
\input{tables/params}

In \cref{tab:params} we report the number of parameters in our model's detection branch, GLASS component and the recognition branch.
During inference, GLASS preforms two main operations: (a) Sampling high-resolution crops and extracting their local features, and (b) Fusing the global features of each detected bounding box with its local feature counterparts.

To minimize the added computational cost, we utilize a light-weight ResNet34 backbone~\cite{focusing_att2017} for extracting the local features.
As described in \cref{sec:GLASS}, we use a block-based attention for fusing the local and global feature maps.
This alleviates much of the fusion computational cost, while still benefiting overall performance, as demonstrated in our experiments.
Overall, the addition of GLASS leads to an inference time increase of roughly 10\%.
The latency increment is similar when incorporating GLASS into Mask TextSpotter v3~\cite{liao2020spotterV3} and ABCNet v2~\cite{liu2021abcnetV2} frameworks, shown in \cref{tab:params}.
GLASS has no effect on the computational cost of the detector, including its mask branch, and the recognizer heads.









\section{Implementation Details}
\label{sec:technical_details}

\subsection{Recognition Head}
\label{sec:recognition}
We follow recent art \cite{liu2021abcnetV2}, and design a light-weight recognition head.
The recognition head consists of 2 convolution layers, a two-layer BiLSTM encoder, and an attention decoder~\cite{clova}.
The input to the first block of the recognizer, the encoder, is the expressive feature map $\mathbf{z}^{\rm fused}$ computed by GLASS component.
The loss used to train the recognition head is the Negative Log-Likelihood (NLL) as in \cite{focusing_att2017}, denoted by $\mathcal{L}_{\rm rec}$.

We note that our GLASS component is modular, and benefits a variety of additional end-to-end recognition branches, including those found in Mask Text Spotter~v3~\cite{liao2020spotterV3}, and ABCNet~v2~\cite{liu2021abcnetV2}.

\subsection{Training and Optimization}

We use a ResNet50 backbone with ImageNet~\cite{russakovsky2015imagenet} pre-trained weights.
Data augmentation includes multiple scales, pixel-level augmentations (color jitter), affine transformations and image cropping. 
In all of our experiments, we set GLASS feature parameters  to ${ C= 256}$, ${ H=8}$, ${ W= 32}$ and ${ k = 8}$.
The recognition head is trained to classify 96 characters, which covers alphabets, numbers, and special characters. 

Recall that the overall loss function $\mathcal{L}$ used for the E2E supervised training is given by
\begin{equation}
    \mathcal{L} =  \mathcal{L}_{\rm rbox} +  \lambda_1  \mathcal{L}_{\rm mask} + \lambda_2  \mathcal{L}_{\rm rec}
    \;,
   \label{eq:total_loss_supp}
\end{equation}
where the mask loss $\mathcal{L}_{\rm mask}$ is identical to Mask RCNN~\cite{he2017mask} and $\mathcal{L}_{\rm rec}$ is the recognition loss, described in \cref{sec:recognition}.
In our experiments, we set $\lambda_1=0.005$, and observe that higher values hurt the E2E text recognition performance.
Also, we empirically set $\lambda_2=2$, and the $\mathcal{L}_{\rm rbox}$ constants are chosen as $\alpha_1=\alpha_2=\alpha_\theta=10$ and $\alpha_3=\alpha_4=5$.
During inference, we resize the longer side of the input image to 1600 for ICDAR15 and ICDAR13 datasets and the shorter side to 1000 for Total-Text and TextOCR datasets.

\section{Further Analysis}
\label{sec:further}

\subsection{Detection Performance}

In this section, we discuss the contribution of GLASS to detection performance.
The recognition head remains unchanged throughout all experiments.
The sole difference are the input features used for recognition during both training and inference.
Here we present in \cref{tab:ablation1_detection}, that contains complementary data to \cref{tab:ablation1}. 

The first observation from \cref{tab:ablation1_detection}, is that we observe a steep performance drop when using only the local branch.
This drop is expected, since when using only the local branch for recognition, the detection backbone can not leverage the supervised recognition as an auxiliary task, which was previously shown to improve detection results in~\cite{qin2019towards,wang2021towards}.

In the same vein, we see marginal differences between the baseline, which attempts to recognize text only using the global features, and the other approaches that combine global and local information.
This is mainly because the back-propagation path and additional supervision to the detection backbone, only occurs via the global feature branch, as can be seen in \cref{fig:framework}.
A simple channel-wise concatenation of global and local features or using GLASS module, rows 3 and 4 respectively, shows only minor gains for detection compared to the baseline.

\input{tables/ablation1_detection}




\subsection{TextOCR Results}
\input{tables/results_textocr}

In \cref{tab:res_textocr}, we report end-to-end text recognition and detection results on TextOCR validation dataset and end-to-end results for TextOCR test dataset.
Test set results are repeated for brevity.  
TextOCR validation and test datasets contain a similar number of images and text instances.
Incorporating GLASS module into the baseline architecture increases the end-to-end  results by a large margin for both validation and test datasets.

\subsection{Qualitative Results}
\begin{figure*}[t]
 \centering
  \includegraphics[width=1.0\linewidth]{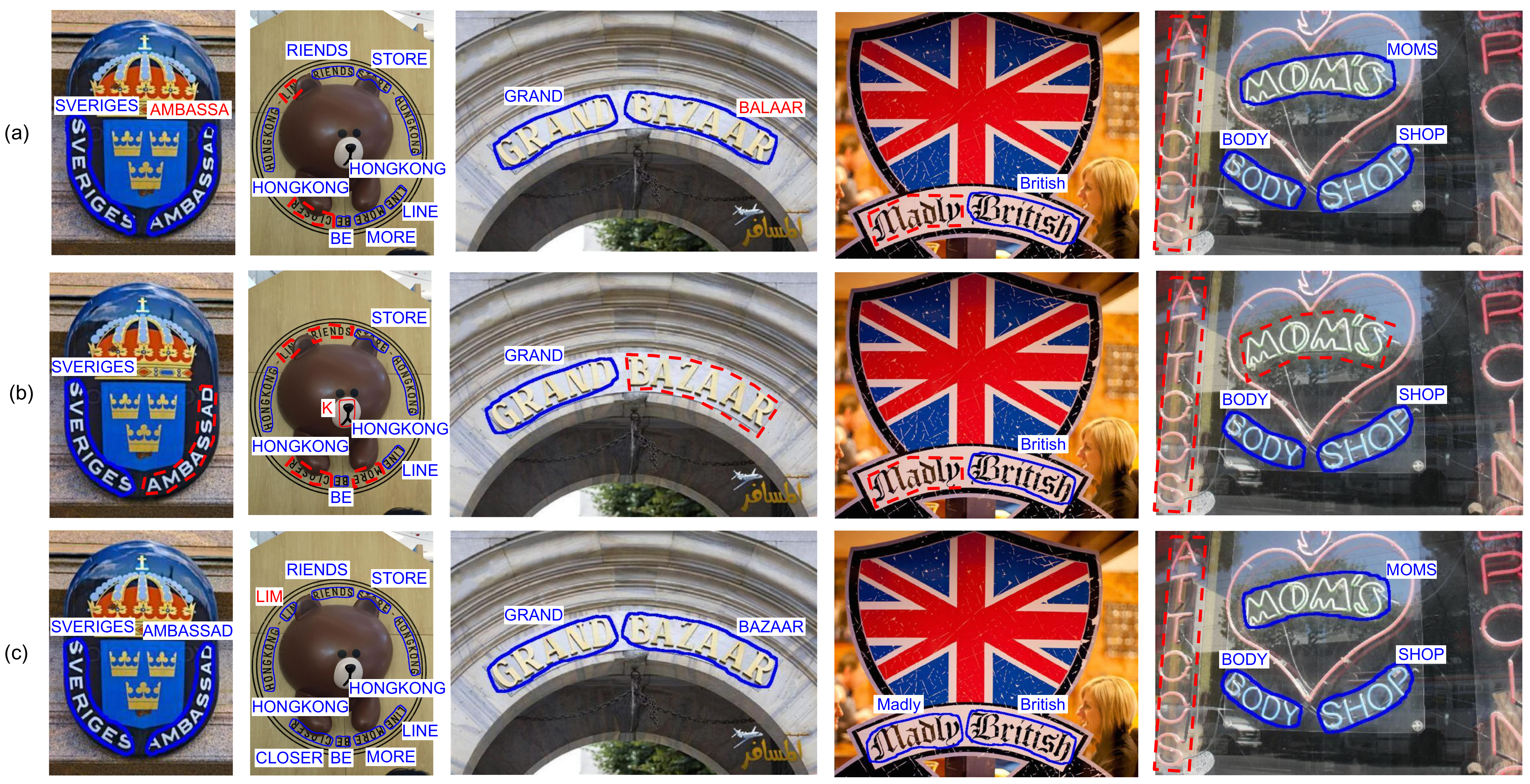}
   \caption{\textbf{Qualitative results on the Total-Text dataset.}
  \textbf{(a)}  Predictions of the baseline experiment, a standard E2E text spotting framework. \textbf{(b)} Predictions of an E2E framework where only the local (crop-level) features are used by the recognizer, and \textbf{(c)} Predictions of our proposed configuration with the GLASS component. Polygons and transcriptions in blue represent correct predictions, and red represents wrong predictions. We observe that an E2E system trained with GLASS is capable of detecting with a higher word recall, and higher recognition accuracy. We recommend enlarging the digital version.}
  \label{fig:totaltext_qualitative_supp}
\end{figure*}
Additional qualitative results from Total-Text are presented in \cref{fig:totaltext_qualitative_supp}.
Our method, \cref{fig:totaltext_qualitative_supp}c, shows high detection and recognition accuracy on curved, rotated, upside down and occluded text and on a variety of fonts.
It can be seen in \cref{fig:totaltext_qualitative_supp}b that using only the crop-level features in the recognizer leads to a regression in both detection and recognition accuracy.
In \cref{fig:totaltext_qualitative_supp}a we present the results of our baseline.
Although the text detection accuracy is qualitatively  high, the lack of crop-level features leads to worse recognition performance compared to our model which uses both local and global features.  

\subsection{In-Depth Scale Performance Analysis}
\label{sec:scale_analysis}

\begin{figure*}[t]
 \centering
  \includegraphics[width=1.0\linewidth]{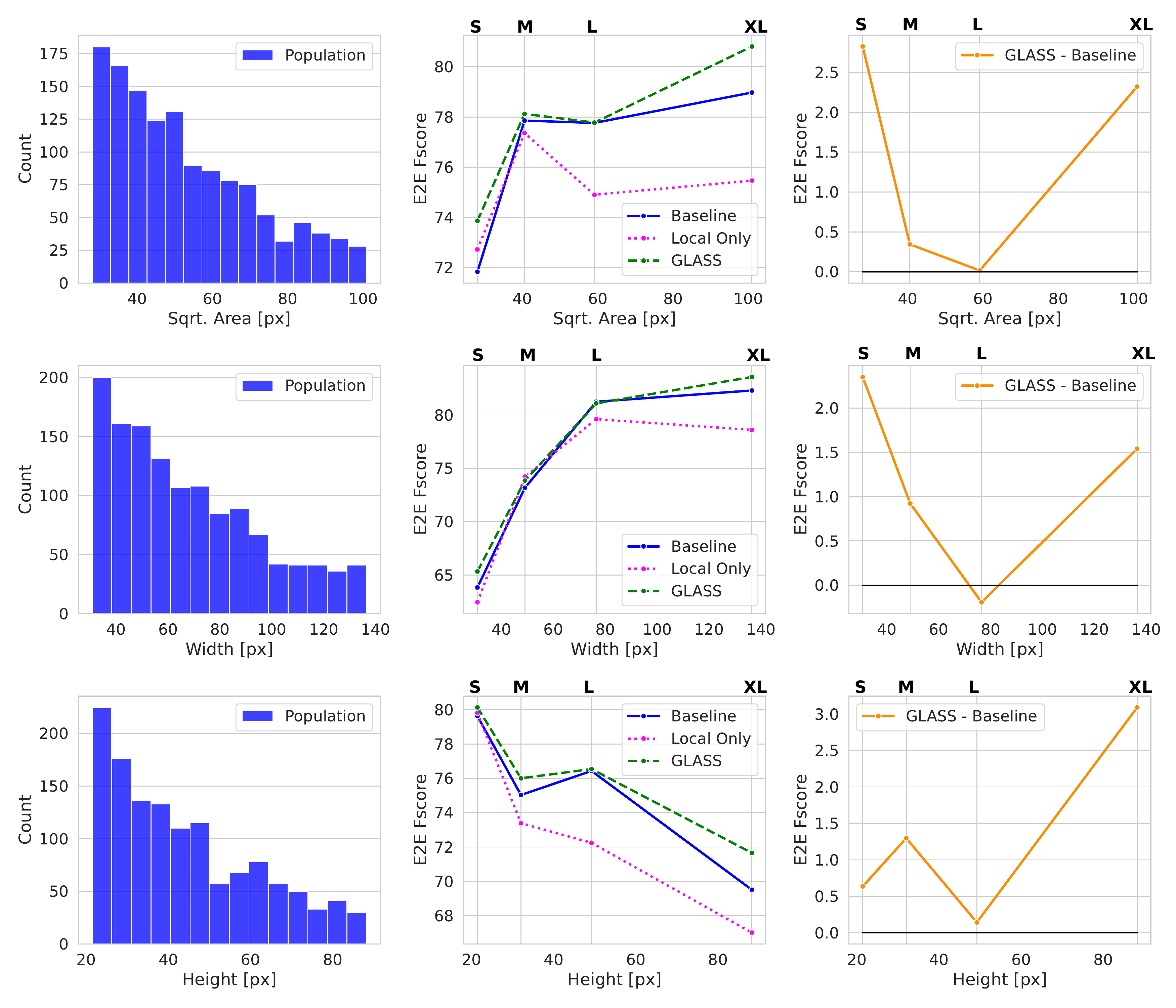}
   \caption{\textbf{In-depth performance analysis for different scales.} Different text instances are over four different size groups S, M, L and XL. 
   We analyze end-to-end recognition performance for 3 different text scale dimensions: (a) Polygon area, (b) Rotated bounding box width and (c) Rotated bounding box height.
   For all different scale properties, GLASS increases performance over the baseline, especially on the extremities of small and large text.
   }
  \label{fig:scale_analysis_supp}
\end{figure*}

We further explore the contribution of GLASS w.r.t. different sized words in an image, and present the performance in relation to additional intrinsic scale properties for word instances in \cref{fig:scale_analysis_supp}. 
The left column in the figure presents the histogram of all word instances and their distribution for both area, width and height. 
The middle column shows the E2E Fscore over four quantiles marked with S, M, L and XL, denoting the relative size groups for the text instances.

We note that regardless of the scale dimension on which we perform an analysis, either area, width or height, we observe the same trend.
GLASS outperforms the baseline by up to 3 percentage points on the edges, measuring the performance on either small or large text instances.

\subsection{Failure Cases}
\begin{figure*}[t]
 \centering
  \includegraphics[width=1.0\linewidth]{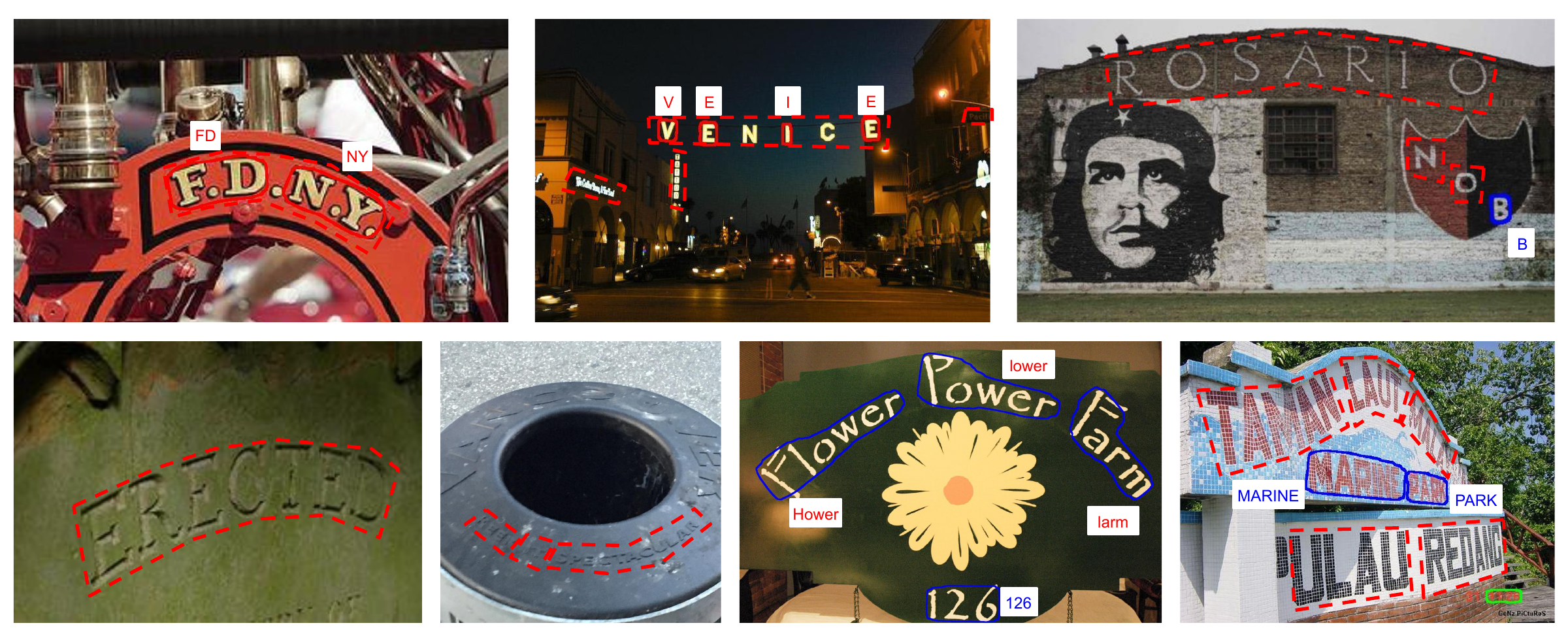}
   \caption{\textbf{Failure cases of our model with GLASS component on the Total-Text dataset.} In the upper images, we see that our model fails to detect text instances with a large space among the characters. In the second row of images, our model struggles to detect and recognize text with irregular font. 
  }
  \label{fig:failure_cases}
\end{figure*}

We show failure cases of our model with GLASS component and our novel orientation loss in \cref{fig:failure_cases}. 
We stress that our model fails to detect text instances with a large space among the characters. 
It may be a result of  our anchor based Mask R-CNN detection branch.
Additionally, we see that our model struggles to detect and recognize text with irregular font.
It may be resolved by training the model on a larger dataset.

%% file: tables/params.tex
\begin{table}[t]
\centering
\caption{\textbf{Model's number of parameters.}  ``FPS'' column states the frames-per-second measured for Total-Text dataset.}
\label{tab:params}
{
\begin{tabular}{l c c cc c c}
\toprule
\multicolumn{1}{c}{\multirow{2}{*}{Method}} &  \multicolumn{1}{c}{Detection} & \multicolumn{1}{c}{\multirow{1}{*}{Recognition}} & \multicolumn{2}{c}{GLASS} & \multicolumn{1}{c}{\multirow{1}{*}{Total \#}} &  \multicolumn{1}{c}{\multirow{2}{*}{FPS}}  \\ 
 &  \multicolumn{1}{c}{Branch} & \multicolumn{1}{c}{\multirow{1}{*}{Head}} & \multicolumn{1}{c}{ResNet34} & \multicolumn{1}{c}{Fusion} & \multicolumn{1}{c}{\multirow{1}{*}{Parameters}} &  \\ 
\midrule
Baseline  & 48.8M & 3.2M &-& -& 52M & 2.7 \\ 
\quad + GLASS& 48.8M & 3.2M & 10.5M & 1.5M & 64M & 3.0 \\ 
\midrule
\midrule
MTSv3~\cite{liao2020spotterV3}  & 41.3M & 4.0M &-& -& 45.3M & 2.3\\ 
\quad + GLASS& 41.3M & 4.0M & 10.5M & 1.5M & 57.3M & 2.6\\ 
\midrule
\midrule
ABCNet v2~\cite{liu2021abcnetV2}  & 44.7M & 3.0M &-& -& 47.7M & 6.0\\ 
\quad + GLASS& 44.7M & 3.0M & 10.5M & 1.5M & 59.7M & 6.5\\ 
\bottomrule
\end{tabular}}
\end{table}


%% file: tables/ablation1_detection.tex
\begin{table}[t]
\centering
\caption{\textbf{Ablation study - Detection.} This table complements paper Table 3 with detection results. ``Fusion Type'' stands the fusion operator used when both feature types are included: channel-wise concatenation and our fusion method. }
\label{tab:ablation1_detection}
\resizebox{\textwidth}{!}{
\begin{tabular}{c c c  c ccc ccc}
\toprule
\multicolumn{1}{c}{} &  \multicolumn{2}{ c }{Features} & \multicolumn{1}{c }{Fusion} & \multicolumn{3}{c }{Total-Text} & \multicolumn{3}{c}{ICDAR15}  \\ 
\cmidrule(lr){2-3} \cmidrule(lr){5-7} \cmidrule(lr){8-10}
\multicolumn{1}{c }{} & \multicolumn{1}{c}{Global} & \multicolumn{1}{ c}{Local}       & \multicolumn{1}{ c }{Type}           & R        & P       & H      & R        & P       & H     \\ 
\midrule
Baseline  & \checkmark &   &        & 85.8 &90.3 &88.0 & 83.2  & 87.9 & 85.5 \\ 
Baseline + Local  &   & \checkmark      & &85.4   & 88.3  &86.8  &  66.6 & 81.4  & 73.3 \\ 
Baseline + Global-Local Concat. & \checkmark & \checkmark & Concat           & \textbf{87.5} & 89.3  &\textbf{88.4} &81.1  & \textbf{89.6} & 85.2  \\ 

\textbf{Baseline + GLASS}&  \checkmark & \checkmark   &  Ours  &     85.5  & \textbf{90.8} & 88.1  & \textbf{84.5} & 86.9 & \textbf{85.7} \\ 
\bottomrule
\end{tabular}}
\end{table}

%% file: tables/results_textocr.tex


\begin{table}[t]
\centering
\caption{\textbf{Results on the TextOCR validation and test datasets.} 
R, P, and H refer to recall, precision and H-mean. 
No lexicon is used.
Our method with GLASS module outperforms Mask TextSpotter v3  on the test set, noting both  approaches were optimized on similar data, including TextOCR train data~\cite{singh2021textocr}.
On TextOCR validation dataset, our method with the GLASS component surpasses the baseline by a large margin for end-to-end recognition and word spotting metrics.}
\label{tab:res_textocr}
\begin{tabular}{l ccc cc  cc}
\toprule
\multicolumn{1}{c  }{\multirow{4}{*}{Method}} & 
\multicolumn{5}{c  }{Validation set} & \multicolumn{2}{c}{Test set}  \\
\cmidrule(lr){2-6}
\cmidrule(lr){7-8}
&\multicolumn{3}{c }{Detection}&  \multicolumn{1}{c}{\multirow{1}{*}{Word}} &  \multicolumn{1}{c }{\multirow{2}{*}{End-to-End}} &
\multicolumn{1}{c}{\multirow{1}{*}{Word}} &  \multicolumn{1}{c}{\multirow{2}{*}{End-to-End}} 
\\ \cmidrule(lr){2-4} 
\multicolumn{1}{c}{}                        & R      & P      & H      & \multicolumn{1}{c}{Spotting}          & \multicolumn{1}{c }{} & \multicolumn{1}{c}{Spotting}          & \multicolumn{1}{c}{}        \\ 
\midrule
MTSv3~\cite{liao2020spotterV3} & -   & -   & -   & - & - & - & 50.8    \\ \hline

{Baseline}       &\textbf{73.3}   &   {82.5}     &  \textbf{77.6}               &    60.6    & {55.6} & 58.0 & 56.8
\\
\textbf{GLASS}       & {71.5}   &   \textbf{84.3}     &  {77.4}      & \textbf{71.3}         &        \textbf{64.8} & \textbf{70.4} &        \textbf{67.1}
\\ \bottomrule
\end{tabular}
\end{table}